\documentclass[sigconf]{acmart}
\renewcommand\footnotetextcopyrightpermission[1]{} 
\usepackage{tikz}
\usepackage{amsmath}
\usepackage{cleveref}
\usepackage{graphicx}

\begin{document}

\date{}

\title{Reinforcement learning guided fuzz testing for a browser's HTML rendering engine}

\author{Martin Sablotny}
\affiliation{%
  \institution{University of Glasgow}
  \country{United Kingdom}}
\author{Bj{\o}rn Sand Jensen}
  \affiliation{%
    \institution{University of Glasgow}
    \country{United Kingdom}}
\author{Jeremy Singer}
\affiliation{%
  \institution{University of Glasgow}
  \country{United Kingdom}}

\begin{abstract}
Generation-based fuzz testing can uncover various bugs and security
vulnerabilities. However, compared to mutation-based fuzz testing, it takes
much longer to develop a well-balanced generator that produces
good test cases and decides where to break the underlying structure to 
exercise new code paths. We propose a novel approach to combine a trained
test case generator deep learning model with a double deep Q-network (DDQN)
for the first time. The DDQN guides test case creation based on
a code coverage signal. Our approach improves the code coverage
performance of the underlying generator model by up to 18.5\% for the Firefox HTML rendering engine
compared to the baseline grammar based fuzzer.
\end{abstract}



\maketitle

\pagestyle{plain}

\section{Introduction}
Many software products are complex systems (e.g.\ browser applications) that interact with externally
provided data (e.g.\ websites). Such data from the outside world must
be validated before it can be processed by the software. If the validation
is not sufficiently rigorous, the data can trigger unintended application behavior, which
might lead to security-related vulnerabilities,
e.g.\ remote code execution or information leaks. Fuzz testing techniques\cite{sutton2007fuzzing}
can be used to uncover the underlying errors leading to vulnerabilities.
However, for complex input data structures, it is time-consuming to develop
test case generators and to fine-tune them to discover code paths that are
difficult to reach.

Major tech companies like Microsoft\cite{ms_onefuzz} and Google\cite{adkins2020building}
not only fuzz test their own products but also provide
fuzzing infrastructure
for external software. By March 2021, Google OSS Fuzz\cite{google_oss_fuzz} 
discovered 23,907 bugs in 316 software projects as highlighted by 
Ding et al.\cite{DBLP:journals/corr/abs-2103-11518}.
During a fuzz test, the software under test encounters various correct and malformed
inputs as produced by the fuzz test generator. As the tests are executed, the execution
of the software is monitored to detect unexpected behavior, like memory corruptions or sudden
termination, and in some cases code coverage.

In general, test case generation for fuzzing
is divided into two categories. First, the \emph{mutation-based} approach
relies on a provided corpus of valid input data, which is then
mutated by replacing subsets of the input with different values.
Secondly, the \emph{generation-based approach} involves the programmatic synthesis of input data based on the underlying data format. It is
essential for test case effectiveness to introduce errors into the generated output to trigger edge cases.

Generally, the generation-based fuzzing approach yields better results more
quickly than the mutation-based approach. However, developing a test 
case generator is time-consuming work. First, the input grammar needs to
be studied and implemented to generate test cases that adhere to the
specification of the input structure.
Secondly, the test case generator needs to be fine-tuned to be able to
uncover unexpected behavior in the program under test by breaking the
rules of the input specification. However, this can lead in many
cases to the program under test rejecting the input all together.


Manually designing and implementing a high-quality, generation-based fuzzer is often a prohibitively time-consuming and expensive process. A partial solution has show itself in the form of machine learning which offers a viable approach for learning fuzzing strategies from existing examples of input data. Such systems have been investigated and demonstrated for complex domains such as PDF documents \cite{godefroid2017learn} and HTML \cite{sablotny2018rnnfuzz}. A major limitation of the aforementioned learned fuzzers is that they are based entirely on the existing input data, hence ignores the response from the software under test (e.g., code coverage and failures) when generating the input data. A potential solution comes in the form of reinforcement learning \cite{russel2010} which provides a coherent framework for learning optimal policies for sequential decision making based on observed rewards (e.g. code coverage). 

Our aim is to design, train and evaluate a complete generation-based fuzzer combining a deep generative model and reinforcement learning with the software under test in the loop. As a case-study, we focus on HTML data and the Firefox HTML rendering engine as
software under test. HTML has over 100 tags that can be further
specialized with the help of multiple different attributes. The standards document that
describes HTML 5 \cite{html5_w3} is over 1200 pages long. Furthermore, an
efficient test case generator needs to strike the right balance between
adhering to the structural rules and breaking them to uncover
code paths were not originally considered during development.

\begin{figure*}[t]
    \centering
    \includegraphics[width=.95\textwidth]{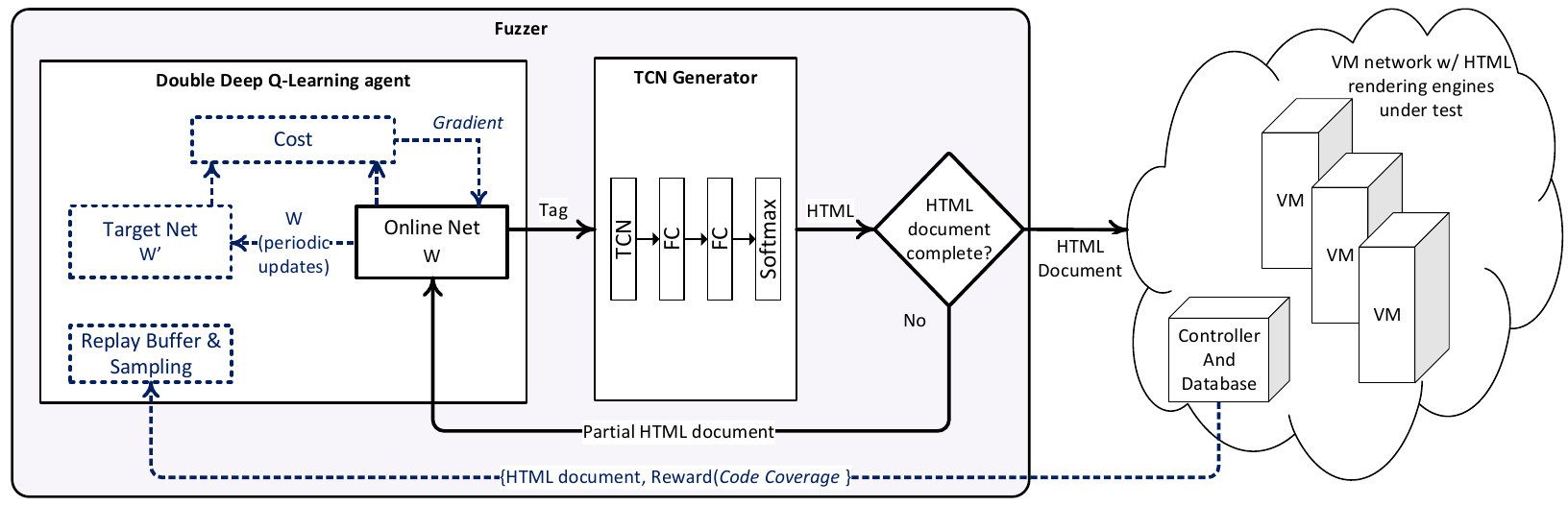}
    \caption{Conceptual overview of the proposed reinforcement learning setup. In training mode (blue/dashed and black elements), the DDQN agent receives the so far generated HTML as state of the world. It decides which HTML tag to insert and then passes the state to the TCN generator model. The generator amends the partial HTML document with newly generated content. This process is repeated until the test case size of 12000 bytes is reached. The code coverage of the test case is then evaluated and used to compute the reward value which is parsed. Once trained, the DDQN agent is still used in the generation process (black elements), but the training (blue/dashed) elements are inactive.}
    \label{exp:RL}
\end{figure*}



We propose to break down the task of learning test case generators informed by code coverage in two parts. The first part is a Temporal Convolutional Network (TCN) which is trained on existing HTML data and used to generate novel HTML test cases conditioned on a specific HTML tag. Secondly, a reinforcement learning model is uses the generator model and feedback from the program execution to lean a policy for increasing the code coverage by selecting the optimal HTML tags. 
Further, we propose the use of a Double Deep Q-Network (DDQN) agent to learn the excepted value of the next action. 
Our combined approach outperforms both (i) a conventional grammar-based test case 
generator and (ii) the TCN generator model alone, in terms of code coverage.
The overall approach is highlighted in \Cref{exp:RL}.
While the approach is demonstrated for fuzz testing HTML rendering engines, the generic approach is applicable in other domains.

Our main contributions are:
\begin{itemize}
    \item A new neural network-based fuzz test case generator for HTML based on the TCN architecture which is able to augment existing fuzzers by increasing their code coverage.
    \item A novel neural network-based fuzzer combining a reinforcement learning DDQN agent with the TCN generator. 
    The agent is informed by code coverage and is trained to maximize code coverage by optimising the generating test cases.
    \item Empirical validation of the performance of the combined DDQN-TCN approach versus
    a default generation-based fuzzer in terms of code coverage in a modern web rendering engine.
\end{itemize}

The rest of the paper is organized as follows. First, \Cref{back} introduces
the necessary background concepts regarding TCNs and DDQN agents. This is followed
by describing the design of the proposed machine learning models in \Cref{design},
which leads to experimental setup in \Cref{exp}. The results are presented in
\Cref{res}. Next, \Cref{disc} discusses the performance of the DDQN agents, 
before \Cref{work} introduces the differences to existing 
research. The second to last part provides an outlook into
future work in \Cref{fut}, and finally, \Cref{con} concludes 
with a summary of our achievements.

\section{Background}
\label{back}
In this section, we start with an introduction to the basic concepts of fuzzing.
This is followed by a review of the concepts behind Temporal Convolutional Networks (TCN) as well
as Double Deep Q-Networks (DDQN) which were used for the experiments.

\subsection{Fuzzing}
Fuzz testing is a software testing approach where the program or function under test
receives synthetic input data that might be malformed. The intention is to trigger code paths
that lead to unintended behavior, like memory corruptions or abrupt program termination.
As mentioned in the introduction, there are two main approaches to fuzz testing one is
to mutate data from an input corpus, and the other is to generate input data programmatically
from scratch. The American Fuzzy Lop (AFL)\cite{afl} is a prominent example of a general-purpose
mutation-based fuzzer because of its performance in uncovering bugs. For generation-based
fuzzing, there is not one particularly good example since these tools are purpose-built
to cover a given input data structure. However, they have the advantage that they have
the input specification hard-wired and can focus on the variable parts of the input
structure. No matter which approach is used, fuzzers need to be configured and fine-tuned
to avoid them getting stuck in a specific code area.

\subsection{Temporal Convolutional Networks}
The concept of a Temporal Convolutional Network (TCN) for sequence modeling was
introduced by Bai et al.\ \cite{bai2018empirical}. The idea is to modify a default
conventional layer as defined by LeCun et al.\cite{lecun1989backpropagation}
with four additional concepts into one layer. First, Bai et al. utilized
one dimensional fully connected convolution from Long et al.\cite{long2015fully}.
This adds padding to the layer's input and makes sure that the input and the output
have the same length. Secondly, causal convolution was added. It ensures that
the model cannot connect at time step $t$ with time steps $t+1$. So, the model
cannot take information from the future into account while computing the present
time step. Thirdly, dilated convolution provides control about the past time steps
that are taken into account while at time step $t$. The effect can be further
amplified by stacking multiple layers with different dilatation rates.
Finally, the use of residual blocks that are basically a filtered shortcut
connection between a block's input and output enables the input data to influence
the output directly.
The advantage of TCNs compared to Recurrent Neural Networks (RNNs) is TCNs can
compute the input in parallel because there is no need to compute time step $t-1$
before computing time step $t$. This allows faster computation and quicker training.
However, the maximum input length is restricted by the available memory, which is
not the case with RNNs. The input length restricts the past time steps the model
has available to compute the next step in the sequence.

The advantages of TCNs over RNNs in this setting are the faster training times
and the on-par performance with RNNs as Bai et al.\cite{bai2018empirical}
have shown. Furthermore, the limitation of the maximum input length does not
tamper with the test case generation in the HTML setting since it is possible
to adjust the sequence length accordingly to at least 12000 characters.

\subsection{Double Deep Q-Networks}
In general, reinforcement learning describes a group of learning algorithms
that receive the state of environment as input and provides an output that
is intended to take the action that leads to the maximum future reward. Deep
reinforcement learning is the subgroup that uses neural networks to make the
prediction. Watkins et al.\cite{watkins1992q} introduced the iterative Q-learning
algorithm. It provides a reinforcement learning agent with the capability
to learn how to maximize the future reward based on the actual state and
an action the agent takes. Mnih et al.\cite{mnih2013playing} introduced
a method using deep neural networks to estimate the Q-function called 
Deep-Q-Network (DQN). They demonstrated
the performance of their new method by playing ATARI games. They also applied
experience replay introduced by Lin et al.\cite{lin1993reinforcement}. 

Hasselt et al.\cite{van2016deep} introduced Double Deep Q-Networks (DDQN) as
an extension to DQNs. The basic idea is to compute a so-called
Q-value for each (state, action)-pair. The Q-value estimates the future average reward of
taking action $a$ in state $s$. The output of the DDQN provides a Q-value
for each action available to the network as a vector. The next action is then
determined by selecting the element with the highest Q-value. Similar to DQNs, 
DDQNs utilize two networks: the target and the online network. The target network periodically
receives the weights of the online network and is used to evaluate the value
of a chosen action in order to compute the loss. This leads to a reduction
of the value overestimation happening in DQNs, and Double Q Networks as highlighted by Haselt et al.\cite{vanhasselt2015deep}.  

To improve the training, an experience replay memory is used. It contains
the actual state $s_t$, the next state $s_{t+1}$, the reward $r_t$ and the action
$a_t$ taken. The elements from the replay memory are chosen based on their priority,
where new memories get the highest priority. After an experience was used for training,
the new priority is based on the computation error as described by Schaul et al.\cite{schaul2015prioritized}.

The main motivation to employ a DDQN is the demonstrated performance with large
input spaces in the past. For example, Mnih et al.\cite{mnih2015human}
achieved superhuman performance with the less optimized DQN agent in
playing ATARI games. The model used the four most recent frames with 84x84 pixels
as input.

\section{Design}
\label{design}

\begin{table}[b]
    \centering
    \begin{tabular}{|l||c|c|c|c|c|c|c|c|c|c|}
    \hline
        \textbf{cfg} & k & $d_1$ & $d_2$ & $d_3$ & $d_4$ & $d_5$ & $d_6$ & $d_7$ & $dense_1$ & $dense_2$ \\
    \hline
        \textbf{01} & 3 & 1 & 2 & 4 & 8 & 16 & 32 & 64 & 512  & 256 \\
    \hline
        \textbf{02} & 3 & 1 & 2 & 4 & 8 & 16 & 32 & 64 & 1024 & 512 \\
    \hline
        \textbf{03} & 5 & 1 & 2 & 4 & 8 & 16 & 32 & -- & 512 & 256 \\
    \hline
        \textbf{04} & 5 & 1 & 2 & 4 & 8 & 16 & 32 & -- & 1024 & 512 \\
    \hline
        \textbf{05} & 9 & 1 & 2 & 4 & 8 & 16 & -- & -- & 512 & 256 \\
    \hline 
        \textbf{06} & 9 & 1 & 2 & 4 & 8 & 16 & -- & -- & 1024 & 512 \\
    \hline 
        \textbf{07} &18 & 1 & 2 & 4 & 8 & -- & -- & -- & 512 & 256 \\
    \hline
        \textbf{08} &18 & 1 & 2 & 4 & 8 & -- & -- & -- & 1024 & 512 \\
    \hline
    \end{tabular}
    \caption{The different configurations evaluated. The kernel size $k$ is fixed over all convolutional layers (max. 7) and the dilation rate $d_i$ is adjusted to cover the whole input sequence (max. 250 characters). Furthermore $dense_i$ provides the number of internal units used.}
    \label{exp:tcn_param}
\end{table}

\begin{table*}[t]
    \centering
    \begin{tabular}{|l||c|c|c||c|c|c||c|c|c||c|c|c||c|c|c|c|c|}
    \hline
        \textbf{config} & $k_1$ & $s_1$ & $f_1$ & $k_2$ & $s_2$ & $f_2$ & $k_3$ & $s_3$ & $f_3$ & $k_4$ & $s_4$ & $f_4$ & $dense_1$ & $dense_2$ & $dense_3$ & $dense_4$ \\
    \hline
        \textbf{01} & 8 & 2 & 8 & 4 & 2 & 16 & 3 & 1 & 32 & 3 & 1 & 64 & 128 & 128 & 128 & 64 \\
    \hline
        \textbf{02} & 8 & 2 & 16 & 4 & 2 & 32 & 3 & 1 & 64 & 3 & 1 & 64 & 128 & 128 & 128 & 64 \\
    \hline
        \textbf{03} & 8 & 2 & 16 & 4 & 2 & 32 & 3 & 1 & 64 & 3 & 1 & 64 & 128 & 128 & 128 & 128 \\
    \hline
        \textbf{04} & 8 & 2 & 32 & 4 & 2 & 64 & 3 & 1 & 64 & 3 & 1 & 64 & 256 & 256 & -- & --\\
    \hline
    \end{tabular}
    \caption{Evaluated hyper-parameter configurations for the DDQN agent. The parameters $k_i$, $s_i$, $f_i$ provide the kernel size, stride and filter dimensions of the CNN layer i respectively. $dense_i$ provides the number of internal units of the dense layer i.}
    \label{exp:tbl_parameters}
\end{table*}

\subsection{TCN test case generator}
The training set was created by converting the HTML data into an integer 
sequence where each integer corresponds to a single character.
The TCN test case generator consists of four modules closely aligned
to the structure proposed by Bai et al.\cite{bai2018empirical}. First, the input
module receives an integer sequence that is translated into an n-dimensional
vector space through an embedding lookup. The embedding layer is also trained, and
the higher dimension output vector increases the distance between
different input characters. Therefore, it enables the model to learn more
quickly.

Secondly, the TCN itself utilises a configurable number of residual blocks.
The residual blocks use different kernel sizes and dilation rates
to maximize the number of steps they could look back at. This was important
to make sure that the first character in the sequence still had a potential
influence on the output. For example, when the first characters describe the
opening HTML tag the model needs this information to form the corresponding
closing tag.

Thirdly, two dense layers summarize the whole TCN output. These two layers
use a configurable amount of internal units. Finally, an output layer provides
a probability distribution over all characters. The next character in the sequence
is then predicted by sampling from the probability distribution for the last
step in the output sequence.

\subsection{DDQN agent and environment}
The DDQN setup consists of the DDQN agent itself and the environment. First,
the DDQN agent receives the generated HTML at the most recent time step as
integer sequence and applies an embedding layer. The embedding layer is 
pre-trained from the TCN generator. This enables faster training.

Secondly, convolutional layers are applied to the embedded input sequence.
The number of layers, kernel sizes and stride steps are configurable. The
DDQN agent can use conventional layers because for the agent there is no
restriction of taking future time steps into account when evaluating the
input sequence. (By way of contrast, the TCN has to be restricted to only evaluate
past time steps when predicting the next step in the sequence.)

Thirdly, there are dense layers with a configurable number of layers as well as
units. These layers summarize the information outputted by the convolutional
layers. Finally, a dense output layer provides the Q-values for the
available action. The actions were defined as inserting a HTML tag (one action
per available tag) and an action allowing the TCN generator on its own that
action does not alter the sequence and just continues the sampling from
the TCN generator.

\section{Experiments}
\label{exp}

The experiments consisted of two phases. The first phase involved
the training and evaluation of TCN generator models to find a consistently
performing set of hyper-parameters. The second phase used a DDQN agent
to maximize the code coverage performance of the TCN generator.

The main baseline for comparison was a set of 16,384 HTML tags generated by
PyFuzz2's \cite{sablotny_pyfuzz2} HTML component. The test cases
were created by inserting 128 HTML tags into a basic HTML template.
The resulting test cases were subdivided into six sets with 128 test
cases each. In order to establish the baseline, the test cases were
used as input for Firefox which was instrumented by DynamoRIO's DrCov
\cite{dynamoRIO} to record the executed basic blocks. The basic blocks were uniquely identified by their
starting offset within Firefox 57.0.1 libxul library. The libxul library
contains the full web engine of Firefox and, therefore, the
parts responsible for rendering HTML. Firefox was executed in headless
mode to skip the overhead in run-time induced by starting the graphical
user interface. The collection itself took place on Ubuntu 18.04.

All models were implemented using Tensorflow \cite{tensorflow2015-whitepaper}
in versions 1.16 and 2.3.0 on Ubuntu Server 18.04 with a single NVIDIA CUDA enabled GPU.

\subsection{TCN based test case generator}
The data set for the TCN training and validation was created by PyFuzz2's 
HTML generation component. In total, 409,000 HTML tags were generated, resulting
in a size of 36MB. The TCN embedding layer provided a 256-dimensional output. 
The number of residual blocks varied from four to seven
layers. The generator's output was a 107-dimensional vector with a scalar value 
for each available character.
It was trained and tested in eight different configurations. \Cref{exp:tcn_param} 
provides an overview of all TCN configurations. Over all configurations,
the settings for kernel size and corresponding dilation rate were chosen to
ensure that at the maximum sequence length the final output sill receives
information from the first character in the input sequence. This resulted in a
starting kernel size of three as recommended for a character based model by
Bai et al.\cite{bai2018empirical} and the largest kernel size tested was
set to 18. The maximum input sequence length was set to 250 characters.


Per configuration, 15 models were trained on three different splits of the
data set. The five different splits were determined randomly but kept constant
for the experiment. During training the validation loss was computed on a validation set
that was not used during training. The validation loss indicates how well the
model performs on inputs that were not encountered during training. 
The model state that achieved the lowest validation loss was used for 
test case generation. Furthermore, early stopping
was applied if five consecutive epochs\footnote{one epoch ends when the training set was seen completely by the model} showed no improvement in validation loss. This allowed to cut down the training time when the model already
reached peak performance.

The aforementioned states with the lowest validation loss were restored
for each model to generate HTML tags for test cases. The starting input
sequence was set to \verb+"<"+ and was used to predict the next character in
the sequence which was concatenated with input sequence and then reused
as new input to the model. This was repeated up to the maximum sequence
length of 250 characters before the input sequence was cut back to the
latest 200 characters. This avoided an additional processing step each
time after the maximum sequence length was reached and therefore increased
test case generation performance. Each model was used to generate 16,384
HTML tags which were placed into 128 test cases with 128 HTML tags each.
The resulting test cases were then executed in Firefox and the
code coverage data was collected.

\subsection{DDQN agent and environment}
The overall implemented environment of the DDQN agent and world is shown in \Cref{exp:RL}. The world
included the evaluation part responsible for executing the test cases and returning
the results. The evaluation part was implemented by provisioning multiple Virtual Machines
(VMs) on an ESXi hypervisor. On these collection VMs, a gRPC network service was running, with
responsibility for receiving the test cases and returning the resulting code
coverage. It executed as the test cases as described above in an instrumented
Firefox instance. The results were returned to a VM with GPU support to perform
the execution and training of the DDQN agent. This central VM also held the experience
replay memory and was connected to a database server used as long-term
storage for the gathered experience to enable offline training and faster hyper-parameter
search. Hyper-parameter search is important because it provides the model with hyper-parameters
which enable the model to perform well. During offline training, the DDQN agent receives the experience
information from the database server directly into the replay memory. This eliminates the
overhead from collecting the code coverage.
Overall, the central VM holds the whole learning environment shown in
\Cref{exp:RL}.

The setup was used to generate and store $2,872,990$ experiences. An experience
contained the state before the DDQN agent action, the state
after the action was performed and the reward. The reward was set to 0 except
when the output size of $12,000$ characters was reached, then the reward was
set to the length of basic blocks discovered by the test cases divided by
the average basic blocks triggered by the underlying TCN generator model.

After the data generation, a hyper-parameter search was conducted. The hyper-parameters
are shown in \Cref{exp:tbl_parameters}. Each configuration was trained
fifteen times, broken down into runs with five different learning rates, each trained
three times. All models were trained on the stored data to enable reproducible
results. The model's performance was evaluated frequently during the training
phase by creating and evaluating four test cases after 20 training steps.
The average code coverage performance of the four test cases was the main indicator
of the model's performance and determined when to store a checkpoint of the model's
state.

The final step was to take the two best performing DDQN agents configurations and
train further 15 models per configuration. The best-performing checkpoint was
then used to generate 128 test cases with a maximum length of $12,000$ characters.
The test cases were then executed to gather the code coverage data and compare the
agent's performance with the underlying TCN generator, and the baseline itself.

\begin{figure}[t]
    \centering
    \includegraphics[width=.45\textwidth]{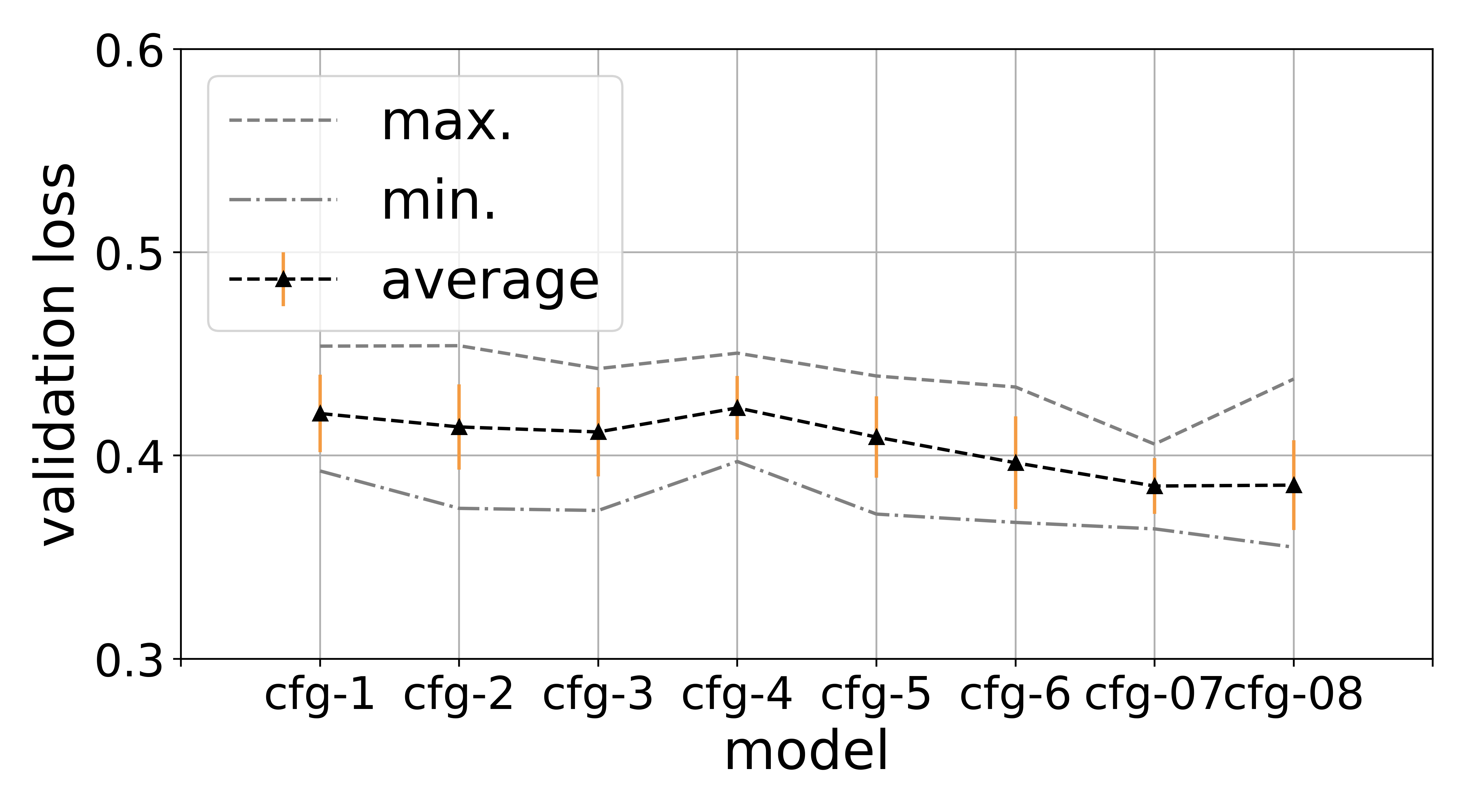}
    \caption{Validation loss for the different TCN model configurations. Showing the minimum, maximum and average loss over all configurations. The average loss data points also show the standard deviation (yellow whiskers).}
    \label{res:fig_tcn_training}
\end{figure}

\section{Evaluation}
\label{res}


\subsection{TCN generator}

The training phase showed a stable validation loss over all configurations
which is supported by a low standard deviation, as highlighted in \Cref{res:fig_tcn_training}. 
The results suggest that a larger kernel size and less layers lead to a lower validation loss.
The TCN configuration number seven (see \Cref{exp:tcn_param}) achieved the lowest validation loss
and standard deviation of $0.385$ and $0.014$ respectively.

\begin{figure*}[t]
    \centering
    \includegraphics[width=.9\textwidth]{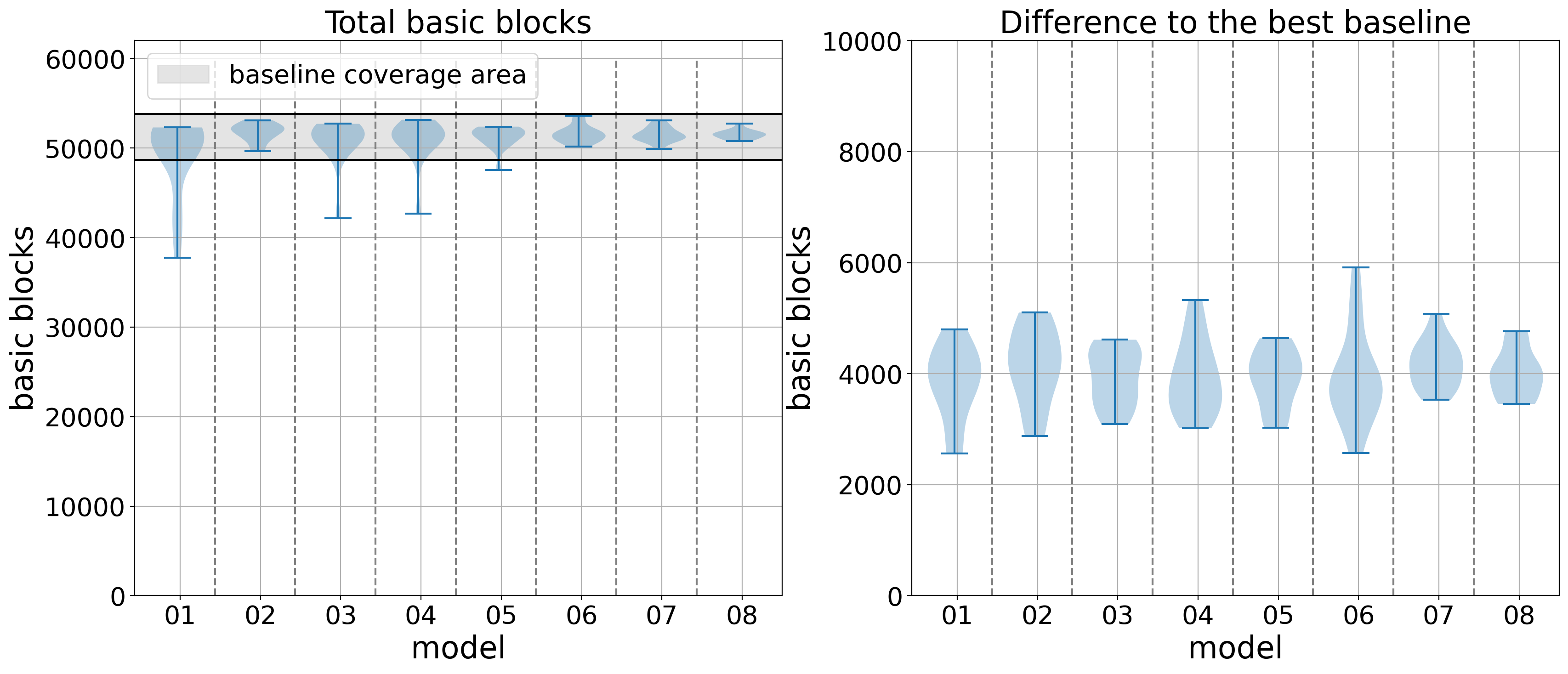}
    \caption{Code coverage performance of the TCN models. (left) total number of unique basic blocks discovered. (right) unique basic blocks not discovered by the dataset}
    \label{res:fig_tcn_perf}
\end{figure*}

The TCN models' code coverage performance was very stable across different configurations, as shown in \Cref{res:fig_tcn_perf}. Furthermore, all trained models were able to achieve a performance similar to the dataset except for 
a few outliers. \Cref{res:fig_tcn_perf} also highlights that the
model configurations that utilize more units in the dense layers were able to perform
better on average. The single best performing TCN model used configuration
six and achieved $53,580$ unique basic blocks, which is only $242$
basic blocks short of the best baseline result. This is a very positive result
and on the same level as the baseline. Furthermore, there are differences in
the actual basic blocks as \Cref{res:fig_tcn_perf}. The difference emphasizes
that the model was actually able to trigger basic blocks that were not triggered
by the best performing baseline set. We chose this model as a generator model for the DDQN agent.

\subsection{DDQN agent}

The training phase of the DDQN agent already showed that it is important to take frequent
snapshots of the model's state and test the performance regularly to ensure 
a good performing configuration is recognized. The average reward for the test
cases fluctuated strongly in between training steps. Furthermore, this also
provides a reason why the lower learning rates perform better on average than
higher learning rates, because the higher learning rates diverge quicker from
well performing states.

\begin{figure*}[t]
    \centering
    \includegraphics[width=.9\textwidth]{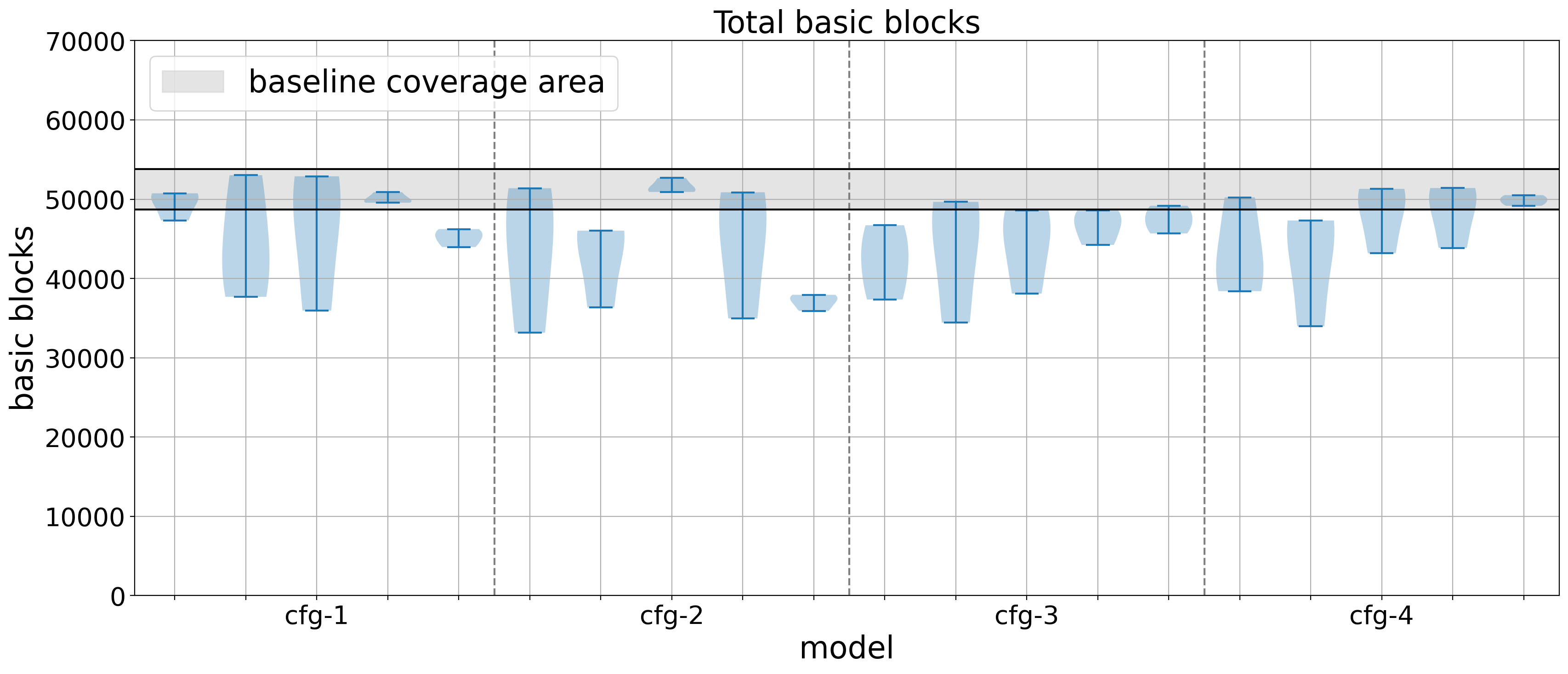}
    \caption{Code coverage performance results of the models trained during hyper-parameter search. Per configuration 5 different learning rates were tested. The violin plots per configuration are ordered by decreasing learning rate.}
    \label{res:fig_hyp_results}
\end{figure*}

We conducted a hyper-parameter search in order to find 
parameters that provide stable and promising results.
The hyper-parameter search provided two promising candidates for
further testing, namely configuration two and four of 
\Cref{exp:tbl_parameters}. The overall results of the hyper-parameter
search can be seen in \Cref{res:fig_hyp_results}. The configuration two 
and four models have the highest on average code coverage performance,
and both also performed best with the learning rate set to $0.000645$.

\begin{figure*}[t]
    \centering
    \includegraphics[width=.9\textwidth]{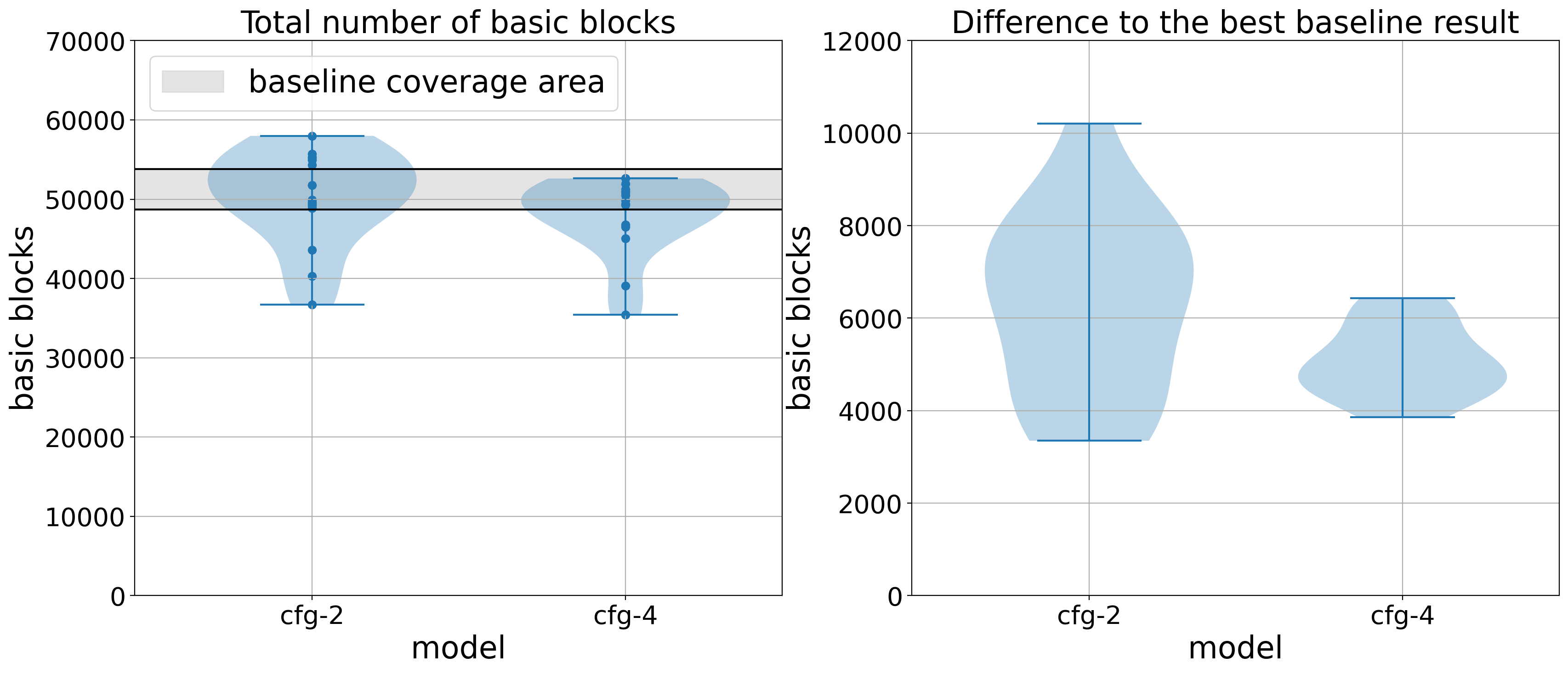}
    \caption{(left) Distribution of code coverage results of the best DDQN configuration compared to the data set baseline. Solid markers show performance of individual models. (right) Difference in uniquely discovered basic blocks between the DDQN agents and the data set.}
    \label{res:fig_perf}
\end{figure*}

\Cref{res:fig_perf} (left) highlights the overall results of the DDQN agents compared
to the baseline results. In total, three configuration two ($\mathcal{C}2$) models were able to outperform the best baseline set
in terms of basic blocks. The best performing DDQN agent achieved $57,993$ 
basic blocks in total that is a 7.7\% performance increase over the best baseline set.
Furthermore, the figure also shows that the majority of $\mathcal{C}2$
DDQN agents either performed close to the maximum data set performance or 
above it. Only three of the fifteen $\mathcal{C}2$ DDQN models performed below the
minimum data set code coverage. The configuration four ($\mathcal{C}4$) models were not able
to outperform the best baseline and achieved a maximum of $52,614$ basic blocks
that is $2.2\%$ below the maximum baseline performance. In total, six out
of the fifteen $\mathcal{C}4$ models were not able to perform above 
the minimum data set performance.


In comparison to the baseline the $\mathcal{C}2$ DDQN agents were able to
discover between $3,349$ and $10,206$ unique basic blocks as highlighted by
\Cref{res:fig_perf} (right). Furthermore, the $\mathcal{C}4$ DDQN agents
discovered between $3,859$ and $6,435$ unique basic blocks. In both 
configurations the DDQN agents were able to improve the overall code coverage by
at least 6\% and in the best case by 18.9\%. 

\begin{figure}[t]
    \centering
    \includegraphics[width=.45\textwidth]{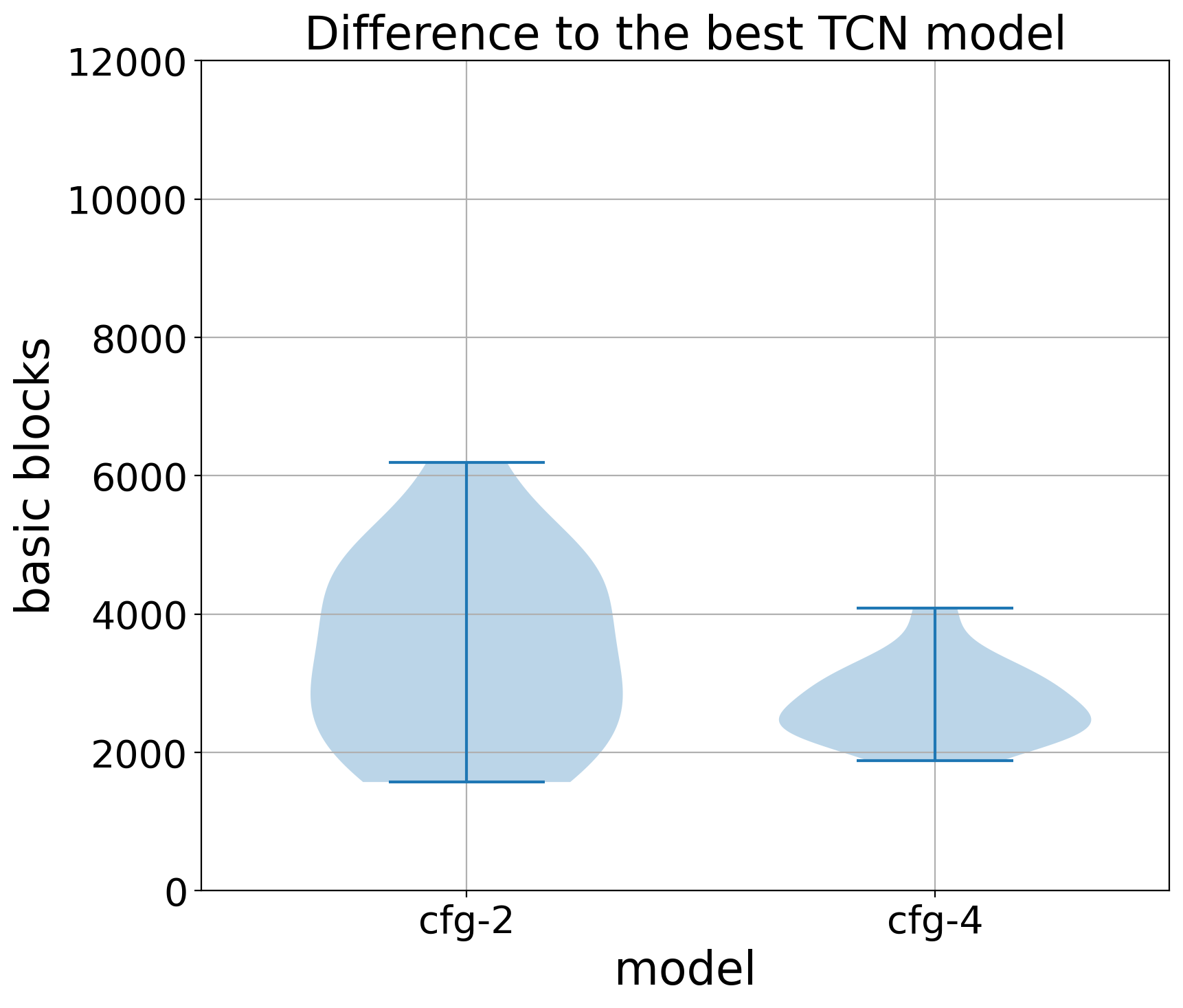}
    \caption{Difference in uniquely discovered basic blocks between the DDQN agents and the used TCN generator model.}
    \label{res:fig_dif_tcn}
\end{figure}

The number of unique basic blocks compared to the underlying TCN generator
unsurprisingly is lower than compared to the baseline as seen in \Cref{res:fig_dif_tcn}.
The $\mathcal{C}2$ DDQN agents achieved $1,573$ to $6,197$ unique basic blocks
and the $\mathcal{C}4$ DDQN agents ranged between $1,886$ and $4,092$ 
This results in an improvement rate of 2.9\% and 11.5\% for worst and best case
respectively.
\section{Discussion}
\label{disc}
The results have shown that it is possible to improve the results of an existing
fuzzer by applying a generative deep learning model and a reinforcement learning
model to the fuzzer.

\begin{figure*}[t]
    \centering
    \includegraphics[width=.9\textwidth]{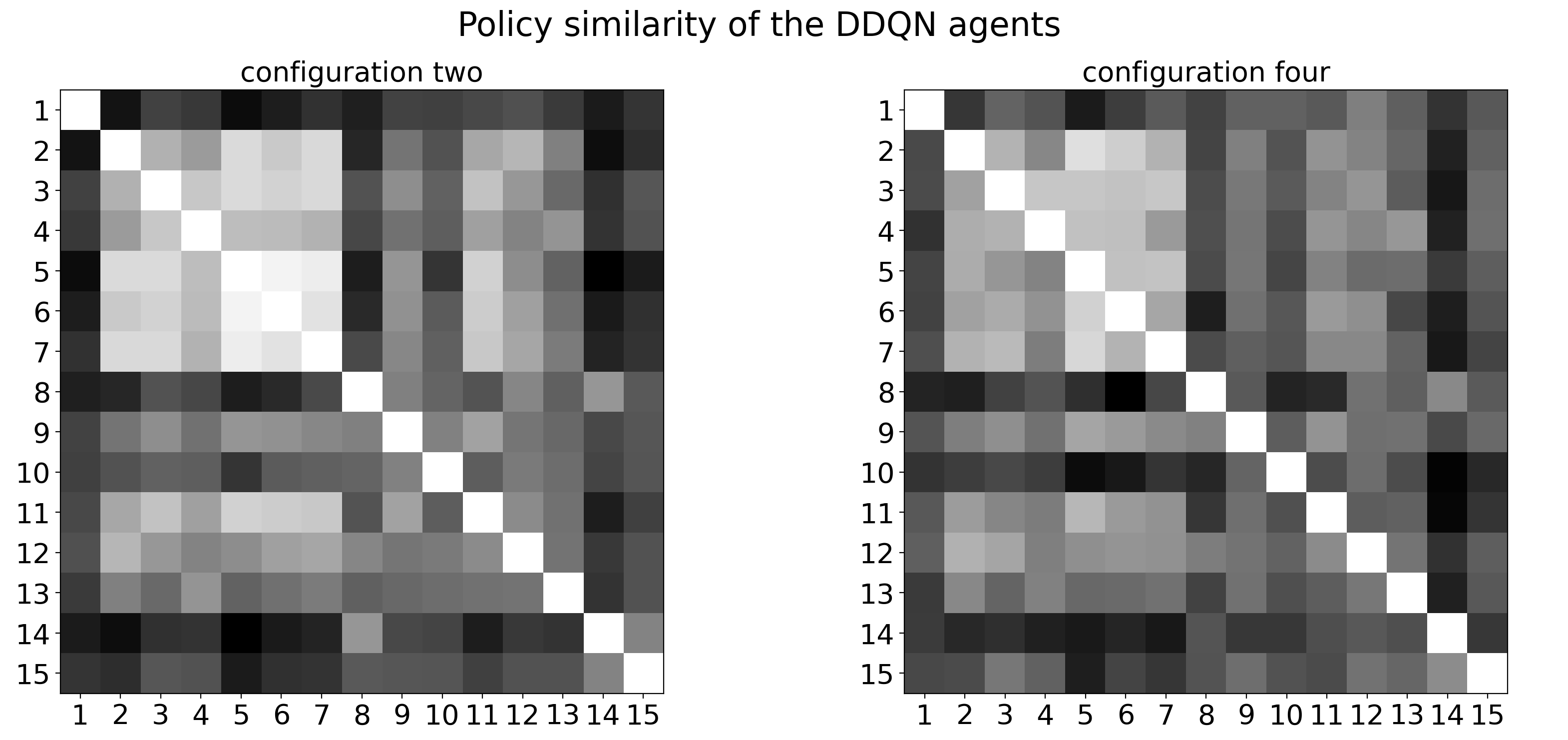}
    \caption{Policy similarity of the 15 trained DDQN agents per configuration. A lighter shade indicates a smaller Kullback Leibler Divergence \cite{kullback1951information} between the policies.}
    \label{disc:distribution}
\end{figure*}

The distribution of chosen actions by the DDQN agents gives interesting insights
into the performance of the model and is highlighted in \Cref{disc:distribution}.
For instance, for the $\mathcal{C}2$ agents, it seems like the low performing
models got stuck with a policy that predicts simple render HTML tags over and
over again whereas the well performing models all have the "input" tag in their
top ten of actions taken. In contrast to that the $\mathcal{C}4$ agents achieve
a similar performance with a more balanced tag distribution. Nonetheless, the low
performing models also show an unbalanced policy with a tendency to 'render only'
tags, like 'a'. Furthermore, computing the 'Kullback Leibler Divergence'\cite{kullback1951information} between the
distributions shows that the well performing action policies have a smaller distance
to each other than to the bad performing ones. For example, the distance between
the training runs six and seven of the $\mathcal{C}2$ models have a distance of
$\approx 2.3211$, whereas the distance between training run six and eight is 
$\approx 9.9995$. The training runs six, seven and eight had a total code coverage
performance of $55,594$, $54,874$ and $36,678$ basic blocks respectively.
The difference in policies is also highlighted by \Cref{disc:distribution} where
the policies of runs two to seven build a similarity block and policy eight is
clearly separated from that block. 

The rewards returned by the VMs indicate a high instability in the training process.
They potentially vary significantly in a few training steps. This effect explains 
why a smaller learning rate worked better during all training runs,
especially with growing model complexity. The initial data collection for the DDQN
agent also indicated that it is beneficial to reuse the TCN embedding weights in
the DDQN agent and disable training of the embedding.

\section{Related Work}
\label{work}
Godefroid et al. \cite{godefroid2017learn} proposed a two layer sequence-to-sequence RNN to
train a PDF-object test case generator with the goal to improve code coverage. They observed a
tension between generating valid test cases and actually inserting errors because the models
reproduce the input well. Therefore they proposed an algorithm called 'SampleFuzz' that inserts
the lowest predicted probability if the highest probability is above a set threshold.
In contrast to their work we researched a different input format namely HTML-tags that
are more structure reliant than PDF-objects. Furthermore, we provided a detailed analysis
of the effects changes in the model architecture have on the output. Finally, we proposed
to improve the code coverage of a generator model by training a DDQN agent to inject HTML tags.

B\"{o}ttinger et al.\cite{bottinger2018deep} researched the application of deep Q-learning on
mutation based fuzzers for PDF files. They used an existing corpus of PDF files to start
with and evaluated two different training approaches. First based on code coverage and
a second based on the standard state-action-reward memory approach.
Whereas they worked on a mutation based approach, we proposed to apply a DDQN agent
on an existing deep learning generator to improve the code coverage of it. Furthermore,
we saw an improvement on the baseline and the generator alone by utilizing a default
replay memory with the state-action-reward approach.

Sablotny et al.\cite{sablotny2018rnnfuzz} proposed a model architecture based on stacked 
Recurrent Neural Networks to generate HTML test cases. Their results indicated that it is
possible to augment a generation based fuzzer with RNNs and increase the
performance of the underlying fuzzer. In contrast to this earlier work we used a TCN based approach
and highlighted that our design is also able to discover areas that were not covered
by the underlying fuzzer. Furthermore, the earlier work did not add a second model to guide the generator
model to increase the code coverage.

\section{Future Work}
\label{fut}
We have demonstrated the advantage of an reinforcement learning-based fuzzer in a specific context, namely HTML and a specific rendering engine. Limiting the test cases to only contain HTML places an upper bound on code coverage of the complete rendering engine. Furthermore, the code coverage guidance by the DDQN is based upon the average achieved code coverage over the evaluation test cases. This does not take into account repeated execution of the same code
areas.

An obvious way to improve the DDQN-based fuzzer would be to base the reward on a more
complex metric. This metric could potentially take into account the repeated execution
of code areas and reduce the reward accordingly over time. In addition,
it would lead to an online training setting where the DDQN is able to adjust
the policy in the real fuzzing environment. So, the DDQN agent would not
be trained upfront but placed into the environment and trained based on
the immediate feedback over time with an adapting reward. This would
punish the DDQN agent with bad rewards when it gets stuck in certain
code areas.

Furthermore, we have opted to break down the system in two parts. An obvious extension would be to combine the TCN generator and DDQN agent into one system that is trainable as a single entity. At the moment the DDQN agent's action set is restricted to deciding which HTML-tags should inserted. This reduces the room the agent has to explore considerably. A combined system could be pre-trained on a specific file format and then
used a code-coverage guided grammar based generator. The combined system
could learn decisions for every position in the generated input and be an online
agent to counter the limitations of stopping the training described earlier.




Another approach that needs to be explored is the potential of transferring
the provided results to other file formats and programming languages. It
might be valuable to combine recent developments in program synthesis, 
Austin et al.\cite{austin2021program} with fuzzing to improve JavaScript
fuzzing, for example.

\section{Conclusion}
\label{con}

We improved the overall code coverage by applying a DDQN agent 
that guided a TCN generator network. The DDQN agent was trained
based on the code coverage performance of generated test cases
to predict the next HTML tag in the test case. Our experiments 
have demonstrated how a TCN model can be used
to generate HTML test cases that discover basic blocks not
discovered by the underlying generation based fuzzer. Furthermore,
the results were improved by utilizing a DDQN agent that guided
the TCN model while generating test cases. 

Overall, the proposed system was able to improve the total code
coverage by up 18.9\% compared to the generation based fuzzer
used to generate the baseline. This highlights that the proposed
system is able to improve an existing fuzzer.


\section*{Acknowledgments}
We gratefully acknowledge the support of NVIDIA Corporation with the provisioning
of the Titan V and Titan Xp graphic cards used for this research.


\bibliographystyle{ACM-Reference-Format}
\bibliography{paper}

\end{document}